\documentclass{article}
\usepackage{iclr2027_conference,times}
\iclrfinalcopy

\usepackage{amsmath,amssymb,amsthm,mathtools}
\usepackage[dvipsnames,table]{xcolor}
\usepackage{array}
\usepackage{graphicx}
\usepackage{float}
\usepackage{booktabs}
\usepackage{multirow}
\usepackage{subcaption}
\usepackage{wrapfig}
\usepackage{enumitem}
\usepackage{microtype}
\usepackage{algorithm}
\usepackage{algorithmic}
\usepackage{listings}
\usepackage{url}
\usepackage{hyperref}

\usepackage[capitalize,noabbrev]{cleveref}

\usepackage{aliascnt}
\newcommand{\sharedthm}[3]{%
  \theoremstyle{#3}%
  \newaliascnt{#1}{theorem}%
  \newtheorem{#1}[#1]{#2}%
  \aliascntresetthe{#1}%
  \crefname{#1}{#2}{#2s}\Crefname{#1}{#2}{#2s}}

\theoremstyle{plain}

\sharedthm{proposition}{Proposition}{plain}
\sharedthm{lemma}{Lemma}{plain}
\sharedthm{corollary}{Corollary}{plain}
\sharedthm{definition}{Definition}{definition}
\sharedthm{remark}{Remark}{remark}
\theoremstyle{plain}

\DeclareMathOperator{\diag}{diag}
\DeclareMathOperator{\blockdiag}{blockdiag}
\DeclareMathOperator*{\CO}{CO}
\newcommand{\Ab}{\bar{A}}
\newcommand{\Ad}{A_{\mathrm{diag}}}
\newcommand{\Ac}{A_{\mathrm{rot}}}
\newcommand{\R}{\mathbb{R}}

\newcommand{\cutsectionup}{\vspace{-2pt}}
\newcommand{\cutsectiondown}{\vspace{-3pt}}
\newcommand{\cutsubsectionup}{\vspace{-2pt}}
\newcommand{\cutsubsectiondown}{\vspace{-3pt}}

\definecolor{myblue}{HTML}{2F3677}
\definecolor{LightBlue}{HTML}{5B8DEF}
\definecolor{VibrantPink}{HTML}{C2185B}
\hypersetup{
    colorlinks=true,
    citecolor=VibrantPink,
    linkcolor=VibrantPink,
    urlcolor=VibrantPink
}
\definecolor{gaingreen}{HTML}{4F9D3C}
\definecolor{stdgray}{gray}{0.45}
\definecolor{codeblue}{rgb}{0.8,0.2,0.2}
\definecolor{codegray}{rgb}{0.5,0.5,0.5}

\newcommand{\boldresultsnote}{Results within 95\% of the best value are written in \textbf{bold}.}
\newcommand{\tabpmstd}[1]{{\color{gray}\fontsize{5.8}{6.6}\selectfont$#1$}}
\newcommand{\mstd}[2]{#1\,\tabpmstd{\pm#2}}
\newcommand{\bestmstd}[2]{\textbf{#1}\,\tabpmstd{\pm#2}}
\newcommand{\bestcell}[2]{{\setlength{\fboxsep}{1pt}\colorbox{LightBlue!32}{\bestmstd{#1}{#2}}}}
\newcommand{\abcell}[1]{{\setlength{\fboxsep}{1pt}\colorbox{LightBlue!32}{\textbf{#1}}}}
\newcommand{\gainup}[1]{\textcolor{gaingreen}{\fontsize{4.6}{5}\selectfont $\uparrow$#1\%}}
\newcommand{\gaindown}[1]{\textcolor{gaingreen}{\fontsize{4.6}{5}\selectfont $\downarrow$#1\%}}
\newcommand{\abgain}[2]{\abcell{#1}\makebox[0pt][l]{\hspace{0.10em}\raisebox{0.12ex}{#2}}}
\newcommand{\lmnum}[1]{\makebox[2.45em][c]{#1}}
\newcommand{\lamgain}[2]{\makebox[2.45em][c]{\abcell{#1}}\makebox[0pt][l]{\hspace{0.10em}\raisebox{0.12ex}{#2}}}

\newcommand{\oursconf}{\begin{tabular}[c]{@{}c@{}}Blockwise\\conformal\end{tabular}}
\newcolumntype{R}{>{\small}r}

\setlist[itemize]{itemsep=1pt, leftmargin=20pt}

\title{Rotate the Loop:\\ Blockwise Conformality for Looped Transformers}

\author{%
Alexey Zemtsov, Albina Klepach, Vadim Lomshakov, Andrei Polubarov,\\
\textbf{Nikita Lyubaykin, Alexander Derevyagin, Vladislav Kurenkov, Alexander Nikulin}\\[0.15em]
{\mdseries\normalfont\footnotesize dunnolab}%
}

\begin{document}

\maketitle

\begin{abstract}
Looped transformers reuse one shared block across many recurrent steps, so depth becomes a quantity
that can be spent at runtime rather than a stack of distinct layers.
Those extra steps help only if a write made on one recurrent step is still readable on later ones,
and that depends on the linear map applied to the residual between recurrent steps.
We show that whether the loop learns to track state is a geometric question about the form of this
map, and we propose \textbf{blockwise conformal} injections: the residual splits into
coordinate-aligned groups and rotates inside each one, so every group has its own timescale while
the geometry stays exact within it.
Our experiments show a performance increase on a demanding form of state tracking - last-token
accuracy when the rest of the prefix is already correct - and on language modelling, where a signal
written earlier has to be retrieved at the end of the sequence.
\end{abstract}

\cutsectionup
\section{Introduction}
\label{sec:intro}
\cutsectiondown

Transformers compute in parallel across a sequence, and that parallelism costs them depth: with a
fixed number of layers they cannot, under standard complexity assumptions, recognize regular
languages or decide graph connectivity
\citep{merrill2022SatAttnTC0,merrill-sabharwal-2023-parallelism,chiang2023tighter,strobl2023transformers}.
These are not exotic failures.
They are the formal shadow of state tracking and of multi-step reasoning
\citep{liu2023transformers,illusion,wei2022chain}, which is why depth - as opposed to width, or a
short chain of thought - has become a resource worth studying on its own \citep{merrill-logdepth}.

One way to buy depth without buying parameters is to loop a shared block
\citep{universal,yang2024looped,huginn}.
A weight-tied transformer unrolled to $\Theta(\log n)$ depth can recognize any regular language and
solve connectivity, and that is also, empirically, the depth training needs \citep{merrill-logdepth}.
Recurrence is the criterion that unrolling turns into: the same block can be applied at test time
beyond the number of steps it was trained on
\citep{huginn,bansal,schwethelm2026much,saunshi2025reasoning,parcae,fprm,lt2}.

The loop still has to converge.
That is a trade-off between a contraction requirement and signal preservation, and weight tying is
what makes the trade-off binding.
The transition operator between recurrent steps has to contract, or the loop destabilizes: that is
the reason for controlling the spectral radius \citep{parcae}, and for residual scaling and Jacobian
regularization \citep{residual-scaling,stars}.
The same map has to keep a signal readable across many recurrent steps, or a partial answer written
on one step is gone by the step that needs it.
Signal preservation, from this view, is stability of the eigenbasis of the transition operator.
States at different recurrent steps stay comparable when the angles between them are preserved, up
to some tolerance, and then the signal is readable along the depth axis.

Angle preservation is what a conformal transformation is.
The same idea is already how length generalization is obtained on the token axis, by a relative
rotation \citep{rope}, how several timescales are kept in a Clockwork RNN
\citep{koutnik2014clockwork}, and how a magnitude is paired with a phase once the recurrence is
written in the complex plane, as in the LRU and in diagonal state-space models
\citep{lru,gu2022s4d}.

These properties are the reason a loop can contract and still be read.
A single shared scale gives every direction the same lifetime, so the state is split into a few
groups, each with its own rate and its own rotation: a complex diagonal whose magnitudes are tied
inside a group.
Existing methods do the opposite.
Within the scope of our derivation, they rescale each coordinate by its own factor, one direction comes to dominate the eigenbasis, and
relative angles are not preserved, thereby demonstrating that 
a nonnegative real spectrum is the regime that cannot express the state-tracking problems we study
\citep{sarrof2024expressive,grazzi2025unlocking}.

Overcoming this structural limitation requires enforcing angle preservation explicitly by constraining the geometry of the transition operator.
In this paper, we bridge this gap by introducing a \textbf{blockwise conformal parameterization} that guarantees stability along the depth axis while preserving the
integrity of the hidden state. By aligning the mathematical requirements of signal preservation with the architecture of looped 
transformers, we enable robust, long-horizon recurrence at test time.

Our contributions are
\begin{itemize}
\item a theoretical characterization of the destabilizing factor of leaving the eigenbasis of the transition
operator unconstrained in the recurrent form of a looped transformer, together with a
parameterization that constrains it and keeps the hidden state readable across any number of
recurrent steps - a pathway toward test-time scaling that is not capped at the training trip count;
\item improved last-token accuracy on state-tracking word problems, where the model must emit the
correct final product given that the prefix is already correct;
\item improved language-model quality when a signal has to reach the end of the sequence,
with lower validation perplexity and higher LAMBADA, so the same linear transition operator helps on text as well
as on the synthetics.
\end{itemize}

\begin{figure}[t]
\centering
\includegraphics[width=\textwidth]{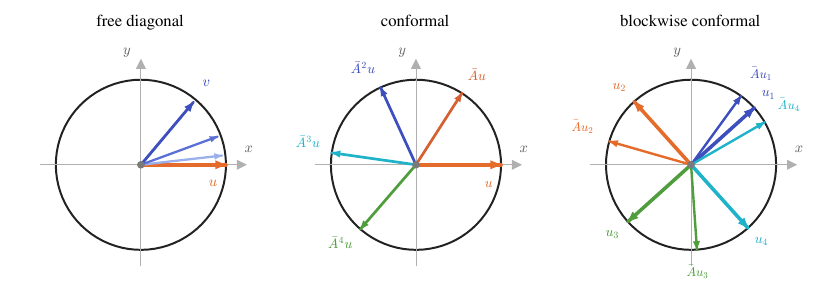}
\caption{\footnotesize
Depiction of how a write moves across recurrent steps, once the transition operator has been applied and the
result renormalized.
The free diagonal stretches each coordinate by its own factor, so two writes that started apart are
dragged onto one axis.
A conformal transition operator is a rotation: the write travels around the circle, and the angle between two
writes stays the angle they were written with.
Blockwise conformal splits the state into a few groups, each turning at its own rate, so the same
rotation still moves the signal, and different groups can age at different speeds.
}
\label{fig:circles}
\end{figure}

\cutsectionup
\section{Related Work}
\label{sec:related}
\cutsectiondown

\textbf{Looped and recurrent-depth transformers.}
Weight tying with a variable trip count goes back to the Universal Transformer \citep{universal} and
has become an axis of test-time compute: $k$ layers looped $L$ times match $kL$ distinct layers on
iterative reasoning and can simulate chain-of-thought steps in latent space
\citep{saunshi2025reasoning}; an adaptive trip count generalizes in length on copying, parity and
addition \citep{fan2025looped}; a $3.5$B recurrent-depth model keeps improving on \textsc{gsm8k} out
to tens of loops \citep{huginn}; and looped models at $1.4$B and $2.6$B trained on trillions of
tokens reach the quality of much larger fixed-depth models \citep{ouro}.
Depth allocated per token rather than per model is a parallel line \citep{mor,bansal}, and on the
theory side $\Theta(\log n)$ looped depth suffices to recognize regular languages and is what
training appears to need \citep{merrill-logdepth}.

\textbf{Parcae.}
\Citet{parcae} treat the recurrent state as a discretized linear recurrence and inject the prelude
embedding at every pass through a diagonal transition operator with entries in $(0,1)$, which bounds the spectral
radius below one and makes long trip counts stable.
Residual scaling, Jacobian regularization and related stabilizers pursue the same end by other means
\citep{residual-scaling,stars,jacobianlens}, and equilibrium treatments take the fixed point as the
object of interest \citep{pathindep,equilibrium,fprm,solvetheloop}.
All of these constrain the \emph{magnitude} of the transition operator.
\Citet{parcae} note that many forms of the injection would work and choose a nonnegative diagonal
because it is simple and cheap; that remark is the opening this paper takes up.

\textbf{Test-time generalization across recurrent steps.}
The Universal Transformer adds a sinusoid of the recurrence index to the state \citep{universal}, and
\citet{xu2025timestep} condition per-loop residual scales and normalization gains on an additive
frequency embedding of the loop index.
An additive index tells the model which recurrent step it is on, and at the same time it restricts
what the model can do: the embedding is tied to the trip counts seen in training, so the loop cannot
freely extrapolate past that range.
On the token axis, rotary embeddings \citep{rope} were the first to put a relative rotation into the
residual geometry; later work generalizes the planes \citep{grape} or accumulates Householder
reflections along the sequence \citep{path}.
Those maps are built per token and applied once.

\textbf{Structured orthogonality and linear recurrences.}
The LRU and diagonal state-space models introduce a complex eigenbasis, with a magnitude and a phase
per plane \citep{lru,gu2022s4d}.
We use that same notion, applied to the recurrence that buys the model test-time compute rather than
to the sequence axis.
\Citet{vorontsov2017orthogonality} show that an orthogonal parameterization improves performance, and
also that a hard constraint is not required: a neighbourhood of the orthogonal family trains even
better.
That softer neighbourhood is the same tendency as a Clockwork RNN
\citep{koutnik2014clockwork,mikolov2015learning} - several timescales instead of one - and it is the
tendency we keep when the state is split into groups.
\Citet{jing2017tunable} build a unitary from layers of $2\times2$ rotations and find that a small
capacity is preferable on copying.
Block-diagonal orthogonal factors appear in parameter-efficient finetuning \citep{boft,gsoft}, on
weight matrices rather than on a recurrent transition operator.

\textbf{Normalization and privileged bases.}
\Citet{slicegpt} establish that an RMSNorm residual stream is exactly invariant to an orthogonal
change of basis, so a rotation applied to the state is invisible to the normalizer that follows it.
\Citet{elhage2023privileged} document that the residual stream nonetheless develops an approximately
axis-aligned privileged basis, attributable to per-coordinate optimizer normalizers.
Together: rotating the residual stream is free as far as the norm is concerned, while the coordinate
frame is not an arbitrary choice - two facts that matter when we place structure in that stream.

\cutsectionup
\section{Motivation of Readability}
\label{sec:theory}
\cutsectiondown

The block of a transformer is nonlinear, and that is where its expressivity lives.
In a loop the same block is applied at every recurrent step, and the cost of that expressivity is
how much of a signal present at the start of the recurrence is still readable at the end.
Between two steps the residual passes through one linear map, the transition operator $\Ab$, and the
block that follows is the same map every time.
Whatever that block computes, it computes as an inner product: attention logits, a linear readout,
the pre-norm rescaling.
The transition operator therefore decides whether those inner products still match the ones the
block would have computed when the signal was written.
Proofs are in \Cref{app:proofs}.

\cutsubsectionup
\subsection{A Readable Operator is a Conformal One}
\label{sec:readability}
\cutsubsectiondown

\begin{definition}[readable]
\label{def:readable}
A matrix $M \in \R^{d\times d}$ is \emph{readable} if there is a $c>0$ with
$\langle Mx, My\rangle = c^{2}\langle x, y\rangle$ for all $x,y \in \R^{d}$.
\label{eq:readable}
\end{definition}

The block compares states by inner products, so a state stays readable when any two vectors can
still be compared after the operator.
The definition says that directly: for every pair $x,y$, the inner product of $Mx$ and $My$ is the
inner product of $x$ and $y$, up to one scale shared by the whole state.
The comparison of $x$ with $y$ is the comparison it was.
An operator that treats directions differently fails the test, because that same pair no longer
presents the same comparison.
That is what a later recurrent step needs if it is to read a state written on an earlier one, and
it is the property the transition operator has to stabilize.

\begin{proposition}[readability is conformality]
\label{prop:read-conf}
$M$ is readable with constant $c$ if and only if $M = cQ$ with $Q^{\!\top}Q = I$.
\end{proposition}

Readability, read strictly, is conformality: the transition operator preserves the angles between
eigenvectors.
We write $\Ac = cQ$ for such a map.

\begin{corollary}[what the block sees]
\label{cor:angles}
If $\Ab = cQ$ then for all $x,y$,
\begin{equation}
\cos\angle(\Ab x,\Ab y) = \cos\angle(x,y),
\qquad
\|\Ab x - \Ab y\| = c\,\|x-y\|,
\qquad
\frac{\Ab x}{\|\Ab x\|} \;=\; Q\,\frac{x}{\|x\|}.
\label{eq:angles}
\end{equation}
\end{corollary}

A pre-norm divides out the shared scale, so the normalized state that reaches the block is a rotated
copy of what was written.
An operator with a free diagonal rescales each coordinate by its own factor.
One direction then dominates the eigenbasis, and a single global normalizer cannot undo that.

\cutsubsectionup
\subsection{Weight Tying Is a Curse for Readability}
\label{sec:compounding}
\cutsubsectiondown

Rotary embeddings \citep{rope} put a rotation on the token axis and apply it once.
On the depth axis the same kind of map is applied to its own output at every recurrent step, and
repeated application of a linear map is power iteration.

\begin{proposition}[iteration collapses a non-conformal operator]
\label{prop:power}
Let $\Ab$ be normal with eigenvalues $\lambda_j$, and let $\mathcal{E}$ be the eigenspace of largest
magnitude.
If $\Ab \in \CO(d)$ then $\|\Ab^{t}u\| = c^{t}\|u\|$ for every $u$, and
$\Ab^{t}u/\|\Ab^{t}u\| = Q^{t}(u/\|u\|)$ moves on the unit sphere without converging.
If instead the largest magnitude is strictly dominant, then
$\operatorname{dist}(\Ab^{t}u/\|\Ab^{t}u\|,\ \mathcal{E}) \to 0$ geometrically
for every $u$ with a nonzero component in $\mathcal{E}$.
\end{proposition}

A transition operator that scales directions differently does not only fade a write.
It turns every write toward one subspace, so writes from different recurrent steps end up pointing
the same way.
A conformal transition operator does not: the direction keeps moving, and the block can still tell
the steps apart.
Readability at one step propagates to every later step, and the distortion between those steps is
estimated by a cosine, one for every pair of depths.
For $u \ne 0$ let
\begin{equation}
G_u[t,s] \;=\; \frac{\langle \Ab^{t}u,\ \Ab^{s}u\rangle}{\|\Ab^{t}u\|\,\|\Ab^{s}u\|},
\qquad 0 \le s,t \le T,
\label{eq:gram}
\end{equation}
the normalized Gram matrix of a single write.
Each entry is how much the geometry of the operator has distorted the state between those two
recurrent steps.

\begin{proposition}[depth invariance]
\label{prop:depthinv}
If $\Ab = cQ$ then $\Ab^{k}$ is readable with constant $c^{k}$ for every $k\ge 1$, and for every
$u\neq 0$, $G_u[t,s] = \langle u, Q^{\,t-s}u\rangle / \|u\|^{2}$,
which depends on $t$ and $s$ only through $t-s$.
\label{eq:toeplitz}
\end{proposition}

\begin{figure}[t]
\centering
\includegraphics[width=\textwidth]{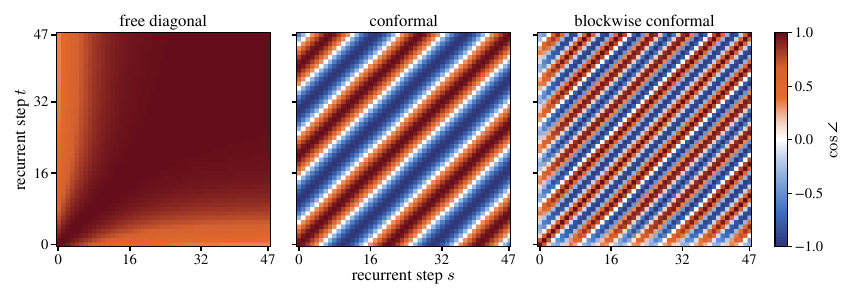}
\caption{\footnotesize
Each panel is the normalized Gram matrix of one write, the cosine of the angle between that write
at recurrent step $t$ and at recurrent step $s$.
Free diagonal: the entries collapse toward one value, so the block cannot tell which recurrent step
the signal came from.
Conformal: the stripes depend only on how many steps apart the two appearances are, so the block
can read the depth at which the signal was written.
Blockwise conformal splits the state into coordinate-aligned groups.
Inside a group the angles stay readable, as in the conformal panel, but the scale is now that
group's own rate, so a write occupying several groups lays several timescales on top of one another.
The finer texture is that granularity: the block can still read the depth of the signal, and each
group keeps it readable for a different number of steps.
}
\label{fig:grams}
\end{figure}

So $G_u$ is constant along its diagonals: two states a fixed number of steps apart present the block
with the same geometry wherever in the loop they sit.
\Cref{fig:grams} illustrates the differences between these operators.

\cutsubsectionup
\subsection{Groups, and the Family They Generate}
\label{sec:groups}
\cutsubsectiondown

One shared scale gives every direction the same lifetime.
The relaxation is to partition $\R^{d} = V_1 \oplus \dots \oplus V_G$ into coordinate subspaces and give
each subspace its own scale,
\begin{equation}
\Ac^{(G)} \;=\; \blockdiag\big(c_1 Q_1,\ \dots,\ c_G Q_G\big),
\qquad c_g > 0,\quad Q_g^{\!\top}Q_g = I_{V_g}.
\label{eq:groups}
\end{equation}
One scale and one orthogonal matrix per group.

\begin{proposition}[blockwise readability]
\label{prop:blockwise}
For $\Ab = \Ac^{(G)}$ as in \eqref{eq:groups} and any $g$, the restriction $\Ab|_{V_g}$ is readable
with constant $c_g$:
$\langle \Ab x, \Ab y\rangle = c_g^{2}\langle x,y\rangle$ for all $x,y \in V_g$.
\label{eq:blockread}
Consequently \Cref{cor:angles}, \Cref{prop:depthinv} and the first half of \Cref{prop:power} hold
verbatim on $V_g$.
For $x \in V_g$ and $y \in V_h$ with $g \neq h$ both sides vanish, and the relative norms of the two
components evolve as $(c_g/c_h)^{t}$.
\end{proposition}

The guarantee holds for any number of groups and any spread of rates, as long as a write lives
inside one group.
What changes across groups is only how long that write lasts: several timescales in one residual,
the Clockwork construction \citep{koutnik2014clockwork} on the depth axis.
$G{=}1$ is one rotation for the whole state; letting every plane keep its own rate frees the
spectrum.

\begin{proposition}[identification]
\label{prop:complex}
Let $Q_g = \blockdiag(R(\theta_i))_{i \in V_g}$ with $R(\theta)$ the planar rotation by $\theta$.
Then \eqref{eq:groups} is unitarily equivalent to the complex diagonal
\begin{equation}
\Ac^{(G)} \;\cong\; \diag\big(c_{g(i)}\,e^{\mathrm{i}\theta_i}\big)_{i=1}^{d/2},
\label{eq:cplx}
\end{equation}
whose eigenvalue \emph{phases} are free per plane and whose \emph{magnitudes} are tied within a group.
At $G = 1$ all magnitudes are tied; at $G = d/2$ they are free.
\end{proposition}

Equation \eqref{eq:cplx} is that family written in the complex eigenbasis: a phase per plane, a
magnitude shared inside a group \citep{lru,gu2022s4d}.
Here $t$ counts recurrent steps.
Setting every phase to zero recovers a nonnegative diagonal, which cannot express the state-tracking
problems we study \citep{sarrof2024expressive,grazzi2025unlocking}.

\cutsectionup
\section{Blockwise Conformal Injections}
\label{sec:method}
\cutsectiondown

A looped transformer runs one weight-tied block $\mathcal{B}$ for $T$ recurrent steps over a residual
state $h_t \in \R^{d}$, re-injecting the prelude embedding $e$ at every step:
\begin{equation}
h_{t+1} \;=\; \mathcal{B}\big(\Ab\, h_t + \Delta\, B e\big),
\qquad t = 0,\dots,T-1 ,
\label{eq:loop}
\end{equation}
with a prelude producing $e$ and a coda reading $h_T$.
The published choice \citep{parcae} discretizes a real diagonal state-space recurrence,
\begin{equation}
\Ad \;=\; \exp\!\big(\!-\!\Delta \odot \exp(a)\big) \;=\; \diag(\alpha_i),
\qquad \alpha_i \in (0,1),
\label{eq:diag}
\end{equation}
with $a \in \R^{d}$ and $\Delta = \mathrm{softplus}(\delta) \in \R^{d}_{>0}$ both learned per coordinate.
The spectrum is real and nonnegative by construction, and the $\alpha_i$ are free of one another.
This section replaces \eqref{eq:diag} with the transition operator \eqref{eq:groups}.
We partition the $d$ coordinates into $G$ contiguous groups of equal width and pair the
coordinates inside each group, giving $P = d/2$ planes with $P/G$ planes per group.
Each group $g$ has a rate pair $(\gamma_g, \delta_g)$ and each plane $i$ has a frequency
$\omega_i$:
\begin{equation}
c_g \;=\; \exp\!\big(\!-\!\mathrm{softplus}(\gamma_g)\,\mathrm{softplus}(\delta_g)\big) \in (0,1),
\qquad
\theta_i \;=\; \omega_i\, \mathrm{softplus}(\delta_{g(i)}),
\label{eq:chart}
\end{equation}
and $\Ab = \blockdiag\big(c_g R(\theta_i)\big)$ as in \eqref{eq:groups}, with $R(\theta)$ the planar
rotation.
The coupling $\theta_i = \omega_i \Delta_g$ in \eqref{eq:chart} is the rule of discretization, read on
a complex pole,
$\exp\!\big(\Delta(-\gamma + \mathrm{i}\omega)\big) = e^{-\gamma\Delta}\big(\cos\Delta\omega + \mathrm{i}\sin\Delta\omega\big)$,
so that decay and phase share one timescale inside one group, and \eqref{eq:diag} is the special
case $\omega \equiv 0$.
Swapping the two changes the structure of $\Ab$ and not the contraction rate training starts from.

$\Ab$ is never materialized.
Because the transition operator is a product of planar rotations, we construct it with the kernel rotary position
embeddings apply to queries and keys \citep{rope}, here applied to the recurrent state on the depth
axis rather than to token positions, so we can still use the RoPE kernel.
Further details are in the implementation of \Cref{app:impl}.

There are no other changes in the architecture.
The block, the prelude, the coda, the tokenizer, the objective, the optimizer, the schedule and the
data stay as they are.
The only edit is the body of the injection module, so every comparison in \Cref{sec:experiments}
differs in the $d \times d$ map applied between recurrent steps.

\cutsectionup
\section{Experiments}
\label{sec:experiments}
\cutsectiondown

\cutsubsectionup
\subsection{Blockwise Conformality Improves State Tracking}
\label{sec:exp-a5}
\cutsubsectiondown

We follow the synthetic state-tracking scenario of \citet{merrill-logdepth}.
The model reads generators $g_1,\dots,g_n$ of a finite group and has to emit every prefix product
$g_1\cdots g_t$.
Training uses the staged length curriculum that paper reports is needed to learn long $A_5$ words,
on those sequences.
We train Parcae looped transformers at width $512$ with Adam at $10^{-4}$, drawing the recurrence
count from a truncated Poisson as in \citet{parcae}.
The baseline is the published learned diagonal \eqref{eq:diag}, trained identically.
Last-token accuracy scores only the full product, the one position that has to survive the entire
loop, and that is the number we report.
Every group-word number is a mean over $10$ seeds with its standard error.
Full configurations are in \Cref{app:repro}.

\begin{table}[t]
  \centering
  \begin{footnotesize}
  \setlength{\tabcolsep}{1.5pt}
  \renewcommand{\arraystretch}{1.05}
  \caption{
  \textbf{State tracking on $A_5$ and $S_5$.}
  Last-token accuracy on word problems of length $n{=}128$, width $512$, trained with the recurrence
  count drawn from a truncated Poisson at $T_{\max}{=}32$.
  $A_5$ is on the left and $S_5$ on the right: the same injections, on a larger group that also
  contains odd permutations.
  Early prefixes resolve within a few steps, so the score is the full product, the one position that
  has to survive the whole loop.
  $T{=}16$ and $T{=}32$ are in-budget evaluations; $T{=}48$ is the same checkpoint at a recurrence
  budget it was never trained on.
  Cells are mean $\pm$ standard error over $10$ seeds.
  $G{=}256$ is one group per rotation plane, the finest chart the width $d{=}512$ affords.
  \boldresultsnote{} Best values are highlighted in \textcolor{LightBlue}{blue}.
  }
  \label{tab:a5}
  \label{tab:s5}
  \begin{tabular}{@{}l@{\hspace{0.7em}}l ccc ccc@{}}
    \toprule
    & & \multicolumn{3}{c}{$A_5$} & \multicolumn{3}{c}{$S_5$} \\
    \cmidrule(lr){3-5}\cmidrule(lr){6-8}
    Injection & $G$ & $T{=}16$ & $T{=}32$ & $T{=}48$ & $T{=}16$ & $T{=}32$ & $T{=}48$ \\
    \midrule
    Diagonal & - & \mstd{0.537}{0.098} & \mstd{0.692}{0.097} & \mstd{0.771}{0.079}
      & \mstd{0.398}{0.096} & \mstd{0.531}{0.092} & \mstd{0.604}{0.084} \\
    \midrule
    \multirow[c]{9}{*}{\oursconf}
      & $1$   & \mstd{0.628}{0.066} & \mstd{0.746}{0.061} & \mstd{0.795}{0.050}
        & \mstd{0.511}{0.071} & \mstd{0.624}{0.065} & \mstd{0.681}{0.055} \\
      & $2$   & \bestcell{0.812}{0.058} & \bestcell{0.894}{0.052} & \bestcell{0.926}{0.040}
        & \bestcell{0.706}{0.061} & \bestcell{0.808}{0.054} & \bestcell{0.837}{0.047} \\
      & $4$   & \mstd{0.481}{0.076} & \mstd{0.573}{0.074} & \mstd{0.632}{0.070}
        & \mstd{0.364}{0.074} & \mstd{0.448}{0.072} & \mstd{0.502}{0.068} \\
      & $8$   & \mstd{0.527}{0.071} & \mstd{0.614}{0.067} & \mstd{0.679}{0.063}
        & \mstd{0.409}{0.069} & \mstd{0.491}{0.065} & \mstd{0.554}{0.061} \\
      & $16$  & \mstd{0.661}{0.057} & \mstd{0.766}{0.053} & \mstd{0.819}{0.045}
        & \mstd{0.542}{0.061} & \mstd{0.647}{0.057} & \mstd{0.693}{0.049} \\
      & $32$  & \mstd{0.603}{0.069} & \mstd{0.701}{0.065} & \mstd{0.757}{0.059}
        & \mstd{0.473}{0.067} & \mstd{0.576}{0.063} & \mstd{0.631}{0.057} \\
      & $64$  & \mstd{0.637}{0.063} & \mstd{0.724}{0.057} & \mstd{0.784}{0.052}
        & \mstd{0.518}{0.061} & \mstd{0.601}{0.056} & \mstd{0.664}{0.051} \\
      & $128$ & \mstd{0.676}{0.049} & \mstd{0.779}{0.043} & \mstd{0.824}{0.041}
        & \mstd{0.556}{0.053} & \mstd{0.661}{0.047} & \mstd{0.709}{0.045} \\
      & $256$ & \mstd{0.654}{0.059} & \mstd{0.741}{0.054} & \mstd{0.791}{0.049}
        & \mstd{0.533}{0.057} & \mstd{0.612}{0.053} & \mstd{0.676}{0.049} \\
    \bottomrule
  \end{tabular}
  \end{footnotesize}
\end{table}

\Cref{tab:a5} is that comparison, with the recurrence count sampled and the decay left free.
At the trained budget $T{=}32$ the published diagonal reaches $0.692$ last-token accuracy and the
blockwise conformal transition operator at $G{=}2$ reaches $0.894$, an interval apart with no overlap.
The separation is already there at $T{=}16$, and it is still there at $T{=}48$, a recurrence budget
none of these checkpoints ever saw.
Both methods improve when given extra recurrent steps at evaluation time, which is the behaviour looped
models are prized for; the conformal method simply starts higher and stays higher.

The knob that matters is the group count, and its useful range is narrow.
$G{=}1$, a single rotation and a single lifetime for the whole state, is already ahead of the
diagonal.
$G{=}2$ is the peak.
Past that the curve turns around: $G{=}4$ and $G{=}8$ fall \emph{below} the diagonal baseline, and
the larger counts recover only partially, with $G{=}256$ - one group per rotation plane, the finest
chart this width allows - ending near $G{=}1$.
We read the shape as the two ingredients pulling in opposite directions.
Phases are present at every $G$ and help throughout, which is why the large-$G$ end still lands near
the top of the diagonal's range.
Tied magnitudes are what a small $G$ buys, and giving each of many groups its own rate hands back
exactly the spread the family exists to remove, so the middle of the sweep is the worst of both.
\Citet{vorontsov2017orthogonality} found an analogous shape on the time axis of an RNN - hard
orthogonality is not optimal, an unconstrained spectrum is worse, a small relaxation beats both -
and that finding is similar to the sweep above.

We also test the same comparison on $S_5$ state tracking, the right half of \Cref{tab:a5}.
The task is harder and the accuracies are lower, and the tendency is the same: $G{=}2$ is the peak,
the middle of the sweep dips, and the gap remains at $T{=}48$.
At the trained budget the diagonal reaches $0.531$ and $G{=}2$ reaches $0.808$.

\cutsubsectionup
\subsection{A Neighbourhood of Conformality Accounts for the Gain}
\label{sec:exp-parts}
\cutsubsectiondown

The same group-word protocol varies the two design choices separately: tie the rates inside
coordinate-aligned groups (\Cref{prop:blockwise}), and make the transition operator a rotation rather than a
rescaling (\Cref{prop:read-conf}).

\begin{table}[t]
  \centering
  \begin{small}
  \setlength{\tabcolsep}{5pt}
  \renewcommand{\arraystretch}{1.08}
  \caption{
  \textbf{Ablating the rotation.}
  Last-token accuracy on the $A_5$ protocol of \Cref{tab:a5} ($n{=}128$, width $512$,
  $T_{\max}{=}32$).
  The grouped-diagonal rows tie the published baseline's rates inside $G$ coordinate-aligned groups
  and leave the transition operator elementwise --- the same lifetimes as \Cref{tab:a5}, without a rotation.
  $T{=}16$ and $T{=}32$ are in-budget evaluations; $T{=}48$ is the same checkpoint at a recurrence
  budget it was never trained on.
  Cells are mean $\pm$ standard error over $10$ seeds.
  Every grouped row is above the free diagonal and below the conformal transition operator at the same $G{=}2$.
  \boldresultsnote{} Best values are highlighted in \textcolor{LightBlue}{blue}.
  }
  \label{tab:diaggroups}
  \begin{tabular}{@{}llccc@{}}
    \toprule
    & & \multicolumn{3}{c}{Last-token accuracy} \\
    \cmidrule(lr){3-5}
    Injection & $G$ & $T{=}16$ & $T{=}32$ & $T{=}48$ \\
    \midrule
    Diagonal & - & \mstd{0.537}{0.098} & \mstd{0.692}{0.097} & \mstd{0.771}{0.079} \\
    \midrule
    \multirow[c]{5}{*}{\begin{tabular}[c]{@{}c@{}}Grouped\\diagonal\end{tabular}}
      & $2$  & \mstd{0.721}{0.051} & \mstd{0.810}{0.036} & \mstd{0.858}{0.033} \\
      & $4$  & \mstd{0.694}{0.054} & \mstd{0.782}{0.040} & \mstd{0.831}{0.037} \\
      & $8$  & \mstd{0.668}{0.057} & \mstd{0.756}{0.044} & \mstd{0.808}{0.040} \\
      & $16$ & \mstd{0.644}{0.061} & \mstd{0.734}{0.048} & \mstd{0.788}{0.043} \\
      & $32$ & \mstd{0.619}{0.066} & \mstd{0.718}{0.052} & \mstd{0.786}{0.047} \\
    \midrule
    \oursconf & $2$
      & \bestcell{0.812}{0.058} & \bestcell{0.894}{0.052} & \bestcell{0.926}{0.040} \\
    \bottomrule
  \end{tabular}
  \end{small}
\end{table}

\begin{table}[t]
  \centering
  \begin{small}
  \setlength{\tabcolsep}{6pt}
  \renewcommand{\arraystretch}{1.08}
  \caption{
  \textbf{Ablating the orthogonal parameterization.}
  Three nested writings of a conformal factor on $A_5$ at $n{=}128$, width $512$,
  $T_{\max}{=}16$, scored against the matched slice of the diagonal baseline.
  All three reach $Q^{\!\top}Q = I$, so the setting isolates
  \emph{which} orthogonal matrices the parameterization can reach - the pairwise chart of
  \eqref{eq:chart} covers a $d/2$-dimensional torus, Cayley fills $SO(d)$ up to a null set, and
  Householder QR reaches both components of $O(d)$.
  A general orthogonal factor already beats the diagonal; the most restricted writing beats it most.
  Cells are mean $\pm$ standard error over $10$ seeds.
  \boldresultsnote{} Best values are highlighted in \textcolor{LightBlue}{blue}.
  }
  \label{tab:chart}
  \begin{tabular}{@{}llc@{}}
    \toprule
    Injection & Chart & Last-token accuracy \\
    \midrule
    Diagonal & Coordinate axes & \mstd{0.408}{0.061} \\
    \midrule
    \oursconf
      & Pairwise
      & \bestcell{0.721}{0.042} \\
    Orthogonal & Householder QR, $O(d)$ & \mstd{0.552}{0.054} \\
    Orthogonal & Cayley, $SO(d)$ & \mstd{0.274}{0.059} \\
    \bottomrule
  \end{tabular}
  \end{small}
\end{table}

\Cref{tab:diaggroups} removes the rotation and keeps the grouping: the baseline's $d$ free rates are
replaced by $G$ rates repeated across coordinate-aligned blocks, with the transition operator still elementwise.
Tying rates alone helps, and helps substantially - $0.810$ at $G{=}2$ and $T{=}32$ against the free
diagonal's $0.692$ - and it prefers the same small group count, with the gain shrinking back
towards the baseline as $G$ grows.
Read together with \Cref{tab:a5} this gives a clean decomposition of the $0.692 \to 0.894$ gap at
$T{=}32$.
Tying magnitudes without rotating recovers $0.118$ of it.
Rotating without tying magnitudes, which is the $G{=}256$ row of \Cref{tab:a5}, recovers $0.049$.
Doing both recovers all $0.202$.
Tying rates is the larger of the two effects and the rotation the smaller, the same ordering holds at
$T{=}16$ and $T{=}48$, and neither ingredient reproduces the full gap alone.
That is what the theory says should happen, since one choice fixes the geometry inside a group and
the other decides how many lifetimes the state gets to have.
The useful point is the small-$G$ end: one shared conformal map is already ahead of a free spectrum,
and a neighbourhood of it, a few groups rather than one lifetime or one rate per plane, is better
than either extreme.

\Cref{tab:chart} varies the second choice in a different direction.
\Cref{prop:read-conf} names a group, $\CO(d)$, and not a way of writing it down, and \Cref{app:discussion}
explains why the theory cannot choose between writings: every claim is invariant to an orthogonal
change of basis.
So we compare three parameterizations that all reach $Q^{\!\top}Q = I$, at a matched spectrum, and
let the optimizer decide.
A general orthogonal factor does help - Householder QR reaches $0.552$ against the diagonal's
$0.408$ - but the restricted pairwise chart of \eqref{eq:chart} reaches $0.721$, well above it,
while the Cayley writing of $SO(d)$ reaches $0.274$ and finishes below the diagonal.
Committing to a fixed eigenbasis and learning $d/2$ angles beats learning the whole rotation, and
learning the whole rotation still beats the published diagonal.
The theory is blind to the basis; the optimizer is not, and the direction of the effect is the one
the privileged-basis account predicts \citep{elhage2023privileged}.

\cutsubsectionup
\subsection{Blockwise Conformality Improves Language Modelling}
\label{sec:exp-lm}
\cutsubsectiondown

The full comparison follows the Parcae training scenario \citep{parcae}, about $80$ tokens per
parameter, and changes only the injection module.
Before that run we chose the group count on a smaller model, $60$M parameters at $20$ tokens per
parameter, the Chinchilla allocation \citep{hoffmann2022training}.
Gradients there go through every recurrent step: full backpropagation through time improves
perplexity over the truncated recipe and does not change which group count wins, and the sweep
selects $G{=}2$ (\Cref{app:lmgroups}).
Three sizes, $60$M, $140$M and $370$M, are then trained once with each transition operator at that
group count.
Full configurations are in \Cref{app:repro}.

\begin{table}[t]
  \centering
  \begin{scriptsize}
  \setlength{\tabcolsep}{3.6pt}
  \renewcommand{\arraystretch}{0.90}
  \caption{
  \textbf{Language-model quality improves with a conformal operator.}
  FineWeb validation perplexity ($780$k tokens) and LAMBADA exact match ($5153$ examples) against the
  number of recurrent steps, at the preferred group count $G{=}2$.
  The gap opens with $T$: at $T{=}1$ the two methods are close; from $T{=}8$ the conformal transition operator is
  ahead by about $15\%$ on perplexity and about $20\%$ on LAMBADA, and the margin holds through
  $T{=}16$, $T{=}32$ and $T{=}48$, in the same direction as the test-time depth sweep of
  \citet{solvetheloop}.
  The arrow at $T{=}48$ is the relative improvement over the parameter-matched diagonal.
  \boldresultsnote{} Best values are highlighted in \textcolor{LightBlue}{blue}.
  }
  \label{tab:lm}
  \begin{sc}
  \begin{tabular}{@{}ll c @{\hspace{2.8em}} c c @{\hspace{2.8em}} c @{}}
    \toprule
    & & \multicolumn{2}{c}{validation ppl $\downarrow$} & \multicolumn{2}{c}{LAMBADA $\uparrow$} \\
    \cmidrule(lr){3-4}\cmidrule(lr){5-6}
    Size & $T$ & ours & diagonal & ours & diagonal \\
    \midrule
    \multirow[c]{5}{*}{$60$M}
      & $1$  & \abcell{28.40} & $29.10$ & \lmnum{0.168} & \abcell{0.174} \\
      & $8$  & \abcell{19.08} & $22.45$ & \abcell{0.276} & $0.230$ \\
      & $16$ & \abcell{18.82} & $22.45$ & \abcell{0.279} & $0.231$ \\
      & $32$ & \abcell{18.70} & $22.45$ & \abcell{0.280} & $0.232$ \\
      & $48$ & \abgain{18.62}{\gaindown{17.1}} & $22.45$ & \lamgain{0.281}{\gainup{21.1}} & $0.232$ \\
    \cmidrule(lr){1-6}
    \multirow[c]{5}{*}{$140$M}
      & $1$  & \abcell{26.48} & $27.04$ & \abcell{0.185} & $0.177$ \\
      & $8$  & \abcell{16.56} & $19.48$ & \abcell{0.320} & $0.267$ \\
      & $16$ & \abcell{16.36} & $19.48$ & \abcell{0.323} & $0.269$ \\
      & $32$ & \abcell{16.24} & $19.47$ & \abcell{0.324} & $0.270$ \\
      & $48$ & \abgain{16.18}{\gaindown{16.9}} & $19.47$ & \lamgain{0.325}{\gainup{19.9}} & $0.271$ \\
    \cmidrule(lr){1-6}
    \multirow[c]{5}{*}{$370$M}
      & $1$  & \abcell{18.72} & $19.39$ & \lmnum{0.228} & \abcell{0.233} \\
      & $8$  & \abcell{12.66} & $14.90$ & \abcell{0.427} & $0.356$ \\
      & $16$ & \abcell{12.51} & $14.90$ & \abcell{0.431} & $0.356$ \\
      & $32$ & \abcell{12.42} & $14.90$ & \abcell{0.433} & $0.357$ \\
      & $48$ & \abgain{12.37}{\gaindown{17.0}} & $14.90$ & \lamgain{0.435}{\gainup{21.8}} & $0.357$ \\
    \bottomrule
  \end{tabular}
  \end{sc}
  \end{scriptsize}
\end{table}

Extra depth only pays when the transition operator keeps what the depth produced.
\Cref{tab:lm} is the three-size family across recurrence budgets, and the clearest result is in the
shape of its columns rather than in any single cell.
At $T{=}1$ the two methods are close: perplexity favours the conformal transition operator by a few percent, and
LAMBADA is a wash - in fact slightly behind at two of the three sizes.
That is the control the theory predicts, since a transition operator applied once cannot compound and
\Cref{prop:power} has nothing to say.
From $T{=}8$ the conformal transition operator is ahead everywhere, by about $15\%$ on perplexity and about $20\%$
on LAMBADA.
Then the two methods stop behaving the same way with depth.
The diagonal's perplexity is flat from $T{=}8$ onwards at every size - $22.45$ at $60$M for all of
$T \in \{8,16,32,48\}$, and the same to within a hundredth at $140$M and $370$M - so past about
eight recurrent steps the extra depth buys it nothing at all.
The conformal transition operator keeps improving through $T{=}48$ at all three sizes, ending about $17\%$ ahead on
perplexity and about $20\%$ ahead on LAMBADA.
This is the practical form of the argument in \Cref{sec:compounding}: test-time depth is only worth
spending if what the earlier steps wrote is still legible to the later ones, and a transition operator that
collapses writes onto a common direction runs out of things to compose.

\cutsubsectionup
\subsection{The Trained Map Has the Predicted Geometry}
\label{sec:exp-geom}
\cutsubsectiondown

We further measure this geometry on the language models trained above.
The Toeplitz defect asks whether two appearances of one state, a fixed number of recurrent steps
apart, still present the block with the same comparison wherever in the loop they sit.
That constancy is what depth invariance makes desirable: extra recurrent steps then do not rewrite
the geometry the block reads.
\Cref{app:defect} defines the defect and states the proposition that a vanishing value certifies
conformality, and that a nonzero value certifies the operator is not conformal.
\Cref{tab:toeplitz} is the measurement.
The conformal method stays within precision error at every width and every depth.
The diagonal method does not: the defect grows with the number of recurrent steps, and it is already
larger at larger width, from $0.06$ at $60$M and $T{=}8$ toward $0.3$ by $T{=}32$.
That growth is the collapse of \Cref{prop:power}, and it meets \Cref{tab:lm} where the diagonal
perplexity stops responding to further recurrent steps.

\providecommand{\sci}[2]{\ensuremath{#1{\cdot}10^{#2}}}
\begin{table}[t]
  \centering
  \begin{footnotesize}
  \setlength{\tabcolsep}{3.5pt}
  \renewcommand{\arraystretch}{1.02}
  \caption{
  \textbf{Toeplitz defect of the trained transition operator.}
  $D(T)$ as in \Cref{app:defect}, averaged over $256$ probe directions.
  Conformal values lie within precision error; diagonal cells are mean $\pm$ sd across probes.
  \boldresultsnote{} Best values are highlighted in \textcolor{LightBlue}{blue}.
  }
  \label{tab:toeplitz}
  \begin{tabular}{@{}llcccc@{}}
    \toprule
    & & \multicolumn{4}{c}{Recurrent steps $T$} \\
    \cmidrule(lr){3-6}
    Size & Operator & $8$ & $16$ & $32$ & $48$ \\
    \midrule
    \multirow[c]{2}{*}{$60$M}
      & Conformal & \abcell{\sci{3.6}{-16}} & \abcell{\sci{3.8}{-16}} & \abcell{\sci{4.1}{-16}} & \abcell{\sci{4.7}{-16}} \\
      & Diagonal & \mstd{0.059}{0.011} & \mstd{0.174}{0.031} & \mstd{0.283}{0.023} & \mstd{0.281}{0.027} \\
    \cmidrule(lr){1-6}
    \multirow[c]{2}{*}{$140$M}
      & Conformal & \abcell{\sci{3.8}{-16}} & \abcell{\sci{4.0}{-16}} & \abcell{\sci{4.4}{-16}} & \abcell{\sci{5.0}{-16}} \\
      & Diagonal & \mstd{0.139}{0.023} & \mstd{0.310}{0.012} & \mstd{0.308}{0.028} & \mstd{0.272}{0.029} \\
    \cmidrule(lr){1-6}
    \multirow[c]{2}{*}{$370$M}
      & Conformal & \abcell{\sci{4.3}{-16}} & \abcell{\sci{4.6}{-16}} & \abcell{\sci{5.0}{-16}} & \abcell{\sci{5.4}{-16}} \\
      & Diagonal & \mstd{0.220}{0.047} & \mstd{0.324}{0.022} & \mstd{0.296}{0.029} & \mstd{0.270}{0.029} \\
    \bottomrule
  \end{tabular}
  \end{footnotesize}
\end{table}

\cutsectionup
\section{Conclusion}
\label{sec:conclusion}
\cutsectiondown

We presented a blockwise conformal map between recurrent steps of a weight-tied transformer.
The need it answers is state tracking on the depth axis, with a potentially very large number of
virtual layers obtained by repeating the same block across more recurrent steps.
Those layers count only while a signal written early is still readable late.
The map between steps has to contract, or the loop does not settle, and it has to keep the angles
between states, or the tied block can no longer compare what it wrote.
Left free, one direction comes to dominate as the same map is applied again and again, and the
comparisons collapse.
Conformality, or a neighbourhood of it, is the practical way through: the state shrinks together,
the angles stay, and a few groups are allowed their own pace, so part of the signal outlasts the
rest.
That is what turns a long unroll into computation rather than a fade.
The gains are on an extreme form of state tracking, the word problem in a finite group, and the
same map also improves language modelling.
The direction worth working on is to spend that depth.
Larger models, run for many more recurrent steps than training currently affords, are where a
readable loop should keep using layers it was never shown.

\cutsectionup
\section*{Reproducibility Statement}
\cutsectiondown

Every claim rests on artifacts we describe.
The construction is a self-contained replacement for one module.
Nothing outside that module differs between the methods of a comparison.
We report ten seeds on every group-word table, retain every seed, and build on publicly available
data and a publicly available codebase.
Full configurations, evaluation protocols and the complete implementation of the transition operator are given in
\Cref{app:impl} and \Cref{app:repro}.
Source code and configurations will be released.

\cutsectionup
\section*{Ethics Statement}
\cutsectiondown

This work studies the linear map between passes of a weight-tied transformer.
The experiments use publicly available corpora and synthetic word
problems over finite groups generated by the authors; no human subjects, no personally identifying
information and no new data collection are involved.

\cutsectionup
\section*{AI Use Statement}
\cutsectiondown

We used generative AI tools, in particular a large language model (LLM), as a writing and coding
assistant. Concretely, we used them to improve the clarity and grammar of the exposition and to
produce the plotting and partly implementation scripts. We have reviewed all AI-assisted text and code and take full
responsibility for the final content of this work.

\bibliography{refs}
\bibliographystyle{iclr2027_conference}

\clearpage
\appendix
\cutsectionup
\section{Theory}
\label{app:discussion}
\cutsectiondown

\cutsubsectionup
\subsection{The Design Space in One Picture}
\cutsubsectiondown

\begin{figure}[H]
\centering
\includegraphics[width=\textwidth]{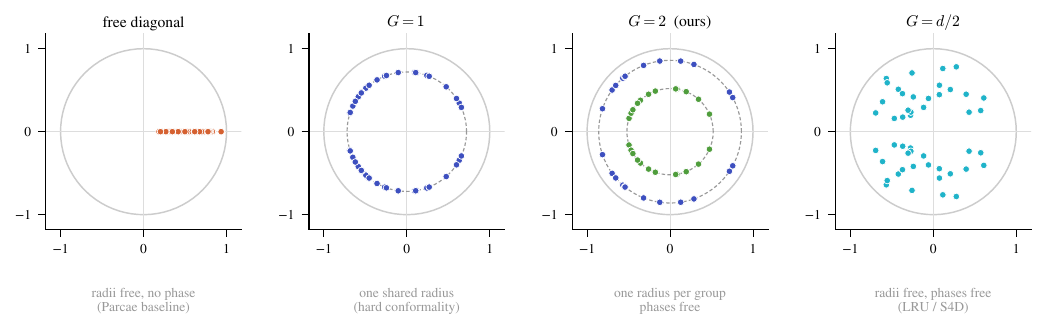}
\caption{\footnotesize
Eigenvalues of the transition operator in the unit disk.
Free diagonal: free radii and no phase.
One shared radius is a single conformal map; one radius per group is the blockwise family;
freeing every radius is a complex diagonal.
}
\label{fig:family}
\end{figure}

\Cref{fig:family} draws the family of \Cref{sec:groups} as a set of eigenvalues in the unit disk, and
it makes the two design choices of \Cref{sec:exp-parts} visible side by side.

Moving from the baseline to our family frees the \emph{phase}, lifting the spectrum off the
nonnegative real axis.
Moving along the family from $G{=}1$ to $G{=}d/2$ frees the \emph{magnitudes}, one group at a time,
until every plane has its own radius and the construction has become an ordinary complex diagonal.
The empirical sweeps in \Cref{tab:a5} and \Cref{tab:lmgroups} traverse this picture from left to
right, and both stop near its left end.

\cutsubsectionup
\subsection{What the Theory Does Not Decide}
\cutsubsectiondown

\begin{proposition}[basis blindness]
\label{prop:basis}
Let $U$ be orthogonal and $\Ab' = U\Ac^{(G)}U^{\!\top}$.
Then $\Ab'$ and $\Ac^{(G)}$ have the same singular values,
and $G_{u}$ computed for $\Ab'$ equals $G_{U^{\!\top}u}$ computed for $\Ac^{(G)}$.
In particular the two transition operators realize the same set of Gram matrices \eqref{eq:gram}
and satisfy \Cref{prop:read-conf}-\Cref{prop:power} identically.
\end{proposition}

The coordinate alignment of \Cref{sec:method} is therefore not a geometric claim but a claim about
which basis the block's parameters can use.
The residual stream of a trained transformer develops an approximately axis-aligned privileged basis,
with basis-aligned outlier dimensions attributable to the per-coordinate normalizers of adaptive
optimizers \citep{elhage2023privileged}.
A dense $U$ spreads each written item over all $d$ coordinates, so the tied readout no longer finds
it where it was put.
\Cref{prop:basis} makes this an empirical question rather than a theoretical one, and
\Cref{tab:chart} treats it as one by comparing three writings of a conformal factor at matched
spectra.

A second freedom is left open in the same way.
\Cref{prop:read-conf} pins down the scale and leaves $Q$ entirely unconstrained, and how to
parameterize $Q$ is an axis with its own literature.
\Citet{jing2017tunable} write a unitary as layers of $2\times2$ rotations with a tunable capacity
$L$, spanning a small subspace at $L=1$ and the full group at $L=N$, and report that the useful
capacity depends on the task.
We take $L = 1$, the most restricted writing available, and \Cref{tab:chart} reports what the larger
ones do: on this task the restricted chart wins, and the least restricted writing we tried finishes
below the baseline.

\cutsubsectionup
\subsection{What Conformality Trades Away}
\label{sec:tradeoff}
\cutsubsectiondown

Nothing in \Cref{sec:theory} says the loop should fail to settle, and it is worth being precise about
what the construction does and does not change.
A free diagonal with $a_i < 1$ and a conformal transition operator with $c < 1$ both contract in \emph{norm};
neither prevents the recurrence from converging.
By \Cref{prop:power} they differ in \emph{direction}.
A non-conformal transition operator additionally converges in direction, onto its dominant eigenspace, and a
conformal one never does.

\cutsubsectionup
\subsection{On Importing Spectral Obstructions}
\cutsubsectiondown

The spectral results we invoke in \Cref{sec:groups} are theorems about the \emph{token}-axis
recurrence of a linear RNN, and we are explicit about what they license.
The negative direction is what rules out the baseline: a nonnegative real spectrum cannot recognize
\textsc{parity} at arbitrary length \citep[Thm.~2]{sarrof2024expressive}, and
\citet[Thm.~1]{grazzi2025unlocking} show an eigenvalue off the nonnegative reals is necessary,
with a non-real one required for counting modulo $m$ \citep[Thm.~2]{grazzi2025unlocking}.
Freeing the phase in \eqref{eq:cplx} removes that obstruction.

The positive direction does not transfer as cleanly, and we do not claim it does.
What those results certify as \emph{sufficient} for composing a non-commutative transition monoid is
a transition that depends on the input, realized as a product of generalized Householders
\citep[Thms.~3-4]{grazzi2025unlocking,siems2025deltaproduct}, and a fixed transition operator iterated $T$ times
is a statement about powers of one matrix.
Our claim on $A_5$ and $S_5$ is empirical: the shape of the transition operator changes what gradient descent
finds, at matched spectra and matched everything else.
Making $\theta$ a function of the token would bring the construction inside the sufficiency results,
costs nothing in the implementation (\Cref{app:impl}), and is the extension we would run next.

\cutsubsectionup
\subsection{Normalization}
\cutsubsectiondown

Pre-normalization is what makes \Cref{cor:angles} operational.
\Citet{slicegpt} show that an RMSNorm residual stream is exactly invariant to an orthogonal change of
basis, which is the property \eqref{eq:angles} needs: a global normalizer removes the shared scale
and commutes with $Q$.
No global normalizer can undo a per-coordinate change of angle, and the angular part is what
\Cref{sec:theory} constrains.

\cutsectionup
\section{The Toeplitz Defect}
\label{app:defect}
\cutsectiondown

For a unit probe $u$ and a trip count $T$, the Toeplitz defect is the root-mean-square distance from
the Gram matrix \eqref{eq:gram} to the nearest matrix that is constant along diagonals:
\begin{equation}
D(T) \;=\; \min_{\tau \in \R^{T+1}} \ \Big(\tfrac{2}{T(T+1)}\!\!\sum_{0 \le s < t \le T}\!\!
\big(G_u[t,s] - \tau_{t-s}\big)^{2}\Big)^{1/2}.
\label{eq:defect}
\end{equation}
In \Cref{sec:exp-geom} the minimum is averaged over $256$ probes drawn uniformly on the sphere.

\begin{proposition}[the Toeplitz defect is a certificate]
\label{prop:converse}
Let $\Ab$ be normal and invertible with eigenvalue magnitudes $r_1 \ge \dots \ge r_d > 0$.
If $r_1 > r_d$ then there is a $u$ for which $G_u$ is not constant along its diagonals.
Hence for normal $\Ab$, $G_u$ is Toeplitz for every $u$ if and only if $\Ab \in \CO(d)$.
\end{proposition}

By \Cref{prop:depthinv} the defect vanishes for a conformal transition operator.
\Cref{prop:converse} is the converse: for a normal operator, a nonzero value certifies that it is
not conformal.
The proof is in \Cref{app:proofs}.

\section{Proofs}
\label{app:proofs}

\subsection*{\Cref{prop:read-conf} (readability is conformality)}

If $M = cQ$ with $Q^{\!\top}Q = I$, then
$\langle Mx,My\rangle = x^{\!\top}M^{\!\top}My = c^{2}x^{\!\top}Q^{\!\top}Qy = c^{2}\langle x,y\rangle$.
Conversely, \eqref{eq:readable} states that $x^{\!\top}(M^{\!\top}M - c^{2}I)y = 0$ for all $x,y \in \R^{d}$.
A bilinear form vanishing on all pairs has zero matrix, so $M^{\!\top}M = c^{2}I$
and $(M/c)^{\!\top}(M/c) = I$. \qed

\subsection*{\Cref{cor:angles} (what the block sees)}

Setting $y = x$ in \eqref{eq:readable} gives $\|Mx\|^{2} = c^{2}\|x\|^{2}$, so $\|Mx\| = c\|x\|$.
Then $\cos\angle(Mx,My) = \langle Mx,My\rangle/(\|Mx\|\|My\|)
= c^{2}\langle x,y\rangle/(c^{2}\|x\|\|y\|) = \cos\angle(x,y)$.
Bilinearity gives $\|Mx-My\|^{2} = \|M(x-y)\|^{2} = c^{2}\|x-y\|^{2}$.
Finally $Mx/\|Mx\| = cQx/(c\|x\|) = Q(x/\|x\|)$. \qed

\subsection*{\Cref{prop:depthinv} (depth invariance)}

$(cQ)^{k} = c^{k}Q^{k}$ and $Q^{k}$ is orthogonal,
so \Cref{prop:read-conf} applies to $\Ab^{k}$ with constant $c^{k}$.
For the Gram matrix,
\begin{equation*}
\langle \Ab^{t}u,\Ab^{s}u\rangle
= c^{t+s}\langle Q^{t}u,Q^{s}u\rangle
= c^{t+s}\langle u,Q^{s-t}u\rangle ,
\qquad
\|\Ab^{t}u\| = c^{t}\|u\| ,
\end{equation*}
using $Q^{\!\top} = Q^{-1}$ in the middle step.
Substituting into \eqref{eq:gram}, the factor $c^{t+s}$ cancels against $c^{t}\cdot c^{s}$,
leaving $\langle u,Q^{s-t}u\rangle/\|u\|^{2}$.
Symmetry in $\pm(t-s)$ holds because $\langle u,Q^{-k}u\rangle = \langle Q^{k}u,u\rangle = \langle u,Q^{k}u\rangle$. \qed

\subsection*{\Cref{prop:converse} (the Toeplitz defect is a certificate)}

Diagonalize the normal matrix $\Ab$ unitarily,
with eigenvalues $\lambda_j = r_j e^{\mathrm{i}\theta_j}$ and orthonormal eigenvectors $v_j$.
Take $u = v_1 + v_d$, so that
$\Ab^{t}u = r_1^{t}e^{\mathrm{i}t\theta_1}v_1 + r_d^{t}e^{\mathrm{i}t\theta_d}v_d$
and $\|\Ab^{t}u\|^{2} = r_1^{2t} + r_d^{2t}$.
At lag one,
\begin{equation*}
G_u[s+1,s] \;=\;
\frac{r_1^{2s+1}e^{\mathrm{i}\theta_1} + r_d^{2s+1}e^{\mathrm{i}\theta_d}}
     {\sqrt{(r_1^{2s+2}+r_d^{2s+2})(r_1^{2s}+r_d^{2s})}} .
\end{equation*}
Divide numerator and denominator by $r_1^{2s+1}$ and write $\beta = r_d/r_1 < 1$:
\begin{equation*}
G_u[s+1,s] \;=\;
\frac{e^{\mathrm{i}\theta_1} + \beta^{2s+1}e^{\mathrm{i}\theta_d}}
     {\sqrt{(1+\beta^{2s+2})(1+\beta^{2s})}} \;\longrightarrow\; e^{\mathrm{i}\theta_1}
\qquad (s \to \infty),
\end{equation*}
a limit of modulus one.
At $s = 0$ the modulus is
$|r_1e^{\mathrm{i}\theta_1}+r_de^{\mathrm{i}\theta_d}| / \sqrt{2(r_1^{2}+r_d^{2})}
\le (r_1+r_d)/\sqrt{2(r_1^{2}+r_d^{2})} < 1$,
where the last inequality is strict precisely because $r_1 \neq r_d$
(equality in $(a+b)^2 \le 2(a^2+b^2)$ requires $a = b$).
So the lag-one diagonal of $G_u$ is not constant.
The reverse implication is \Cref{prop:depthinv},
and a normal $\Ab$ with $r_1 = r_d = c$ satisfies $\Ab^{\!\top}\Ab = c^{2}I$, hence lies in $\CO(d)$. \qed

\subsection*{\Cref{prop:power} (iteration collapses a non-conformal operator)}

The conformal case is \Cref{cor:angles} applied to $\Ab^{t} = c^{t}Q^{t}$.
For the second claim, expand $u = \sum_j a_j v_j$ in the eigenbasis of $\Ab$ and factor out $\lambda_1^{t}$:
\begin{equation*}
\Ab^{t}u \;=\; \lambda_1^{t}\Big(
\sum_{\lambda_j \in \mathcal{E}} a_j v_j
\;+\; \sum_{\lambda_j \notin \mathcal{E}} a_j (\lambda_j/\lambda_1)^{t} v_j \Big).
\end{equation*}
Every ratio in the second sum has modulus $\beta_j = |\lambda_j|/r_1 < 1$,
so that sum vanishes geometrically at rate $\max_{j \notin \mathcal{E}}\beta_j$,
while the first sum is a fixed nonzero vector of $\mathcal{E}$ by assumption.
Normalizing removes the prefactor $\lambda_1^{t}$ up to a unit scalar. \qed

\subsection*{\Cref{prop:blockwise} (blockwise readability)}

$\Ab$ preserves each coordinate subspace $V_g$ and acts on it as $c_gQ_g$ with $Q_g$ orthogonal on $V_g$,
so \Cref{prop:read-conf} applies verbatim on $V_g$ and gives \eqref{eq:blockread}.
Powers act blockwise, $(c_gQ_g)^{t} = c_g^{t}Q_g^{t}$,
so \Cref{prop:depthinv} and the norm identity of \Cref{prop:power} transfer to $V_g$ unchanged.
Distinct $V_g$ are orthogonal as coordinate subspaces and each is $\Ab$-invariant,
so for $x \in V_g$, $y \in V_h$ with $g \neq h$ we have $\Ab x \in V_g$, $\Ab y \in V_h$
and both $\langle x,y\rangle$ and $\langle \Ab x,\Ab y\rangle$ vanish.
The stated ratio follows from $\|\Ab^{t}x\| = c_g^{t}\|x\|$ and $\|\Ab^{t}y\| = c_h^{t}\|y\|$. \qed

\subsection*{\Cref{prop:complex} (identification)}

The planar rotation $R(\theta)$ has eigenvalues $e^{\pm\mathrm{i}\theta}$
with eigenvectors $(1,\mp\mathrm{i})^{\!\top}/\sqrt2$,
so the $2\times2$ change of basis $W = \frac{1}{\sqrt2}\begin{psmallmatrix}1&1\\-\mathrm{i}&\mathrm{i}\end{psmallmatrix}$
is unitary and diagonalizes it.
Applying $W$ blockwise on each coordinate pair diagonalizes every $Q_g$ simultaneously,
and multiplying block $g$ by the scalar $c_g$ multiplies both of its eigenvalues by $c_g$,
which gives \eqref{eq:cplx}.
At $G=1$ there is a single $c$, so all magnitudes coincide;
at $G=d/2$ each plane is its own group and the magnitudes are unconstrained. \qed

\subsection*{\Cref{prop:basis} (basis blindness)}

$\Ab'^{\!\top}\Ab' = U(\Ac^{(G)})^{\!\top}\Ac^{(G)}U^{\!\top}$ is a similarity by an orthogonal matrix,
so it has the same eigenvalues as $(\Ac^{(G)})^{\!\top}\Ac^{(G)}$
and $\Ab'$ has the same singular values as $\Ac^{(G)}$.
Since $(\Ab')^{t} = U(\Ac^{(G)})^{t}U^{\!\top}$ and $U$ preserves inner products and norms,
\begin{equation*}
\frac{\langle (\Ab')^{t}u,(\Ab')^{s}u\rangle}{\|(\Ab')^{t}u\|\|(\Ab')^{s}u\|}
=
\frac{\langle (\Ac^{(G)})^{t}U^{\!\top}u,(\Ac^{(G)})^{s}U^{\!\top}u\rangle}
     {\|(\Ac^{(G)})^{t}U^{\!\top}u\|\|(\Ac^{(G)})^{s}U^{\!\top}u\|},
\end{equation*}
which is $G_{U^{\!\top}u}$ for $\Ac^{(G)}$.
As $u$ ranges over $\R^{d}\setminus\{0\}$ so does $U^{\!\top}u$, so the two sets of Gram matrices coincide. \qed

\cutsectionup
\section{Implementation}
\label{app:impl}
\cutsectiondown

\Cref{alg:op} is the transition operator as run.

\begin{algorithm}[t]
\caption{The transition operator, as run.}
\label{alg:op}
\lstset{basicstyle=\ttfamily\scriptsize,aboveskip=2pt,belowskip=2pt,xleftmargin=0pt}
\begin{lstlisting}
# --- inside the loop: x <- Block(A_bar x + dt * B e), one tied block per pass ---
def core_block_forward(self, x, input_embeds, freqs_cis, mask, step, total_steps, **kw):
    x = self.transformer.adapter(x, input_embeds)      # the transition operator, below
    for block in self.active_blocks:                   # the tied recurrent block
        x = block(x, freqs_cis, mask, **kw)
    return x

def forward(self, x_t, e):                             # the injection module
    return self._apply_A(x_t, self._decay()) + self._dt() * (e @ self.B.T)

def _apply_A(self, x, decay):
    return self._rotate(x) * decay                     # ours: rotate, then group scale

# --- rate and angle: c_g = exp(-softplus(gamma_g) softplus(dt_g)), theta_i = omega_i dt_g ---
def _group_expand(self, v):                            # (G,) -> (d,), one value per group
    return v.repeat_interleave(2 * self.pairs_per_group)

def _decay(self):
    gam, dtg = F.softplus(self.gamma_log), F.softplus(self.rot_dt_log)          # (G,), (G,)
    return self._group_expand(torch.exp(-gam * dtg))   # one modulus per group

def _theta(self):
    dt_g = F.softplus(self.rot_dt_log)                 # decay and phase share one timescale
    return self.rot_omega * dt_g.repeat_interleave(self.pairs_per_group)

# --- the transition operator itself: blockdiag(R(theta_1), ..., R(theta_P)) as a pairwise complex multiply ---
def _rotate(self, x):
    theta = self._theta().to(x.dtype)
    cos, sin = torch.cos(theta), torch.sin(theta)
    xr = x.reshape(*x.shape[:-1], self.rot_pairs, 2)
    x0, x1 = xr[..., 0], xr[..., 1]
    return torch.stack([x0 * cos - x1 * sin,
                        x0 * sin + x1 * cos], dim=-1).reshape(x.shape)
\end{lstlisting}
\end{algorithm}

\textbf{Parameter counts.}
Implemented on its own, the transition operator uses $d/2 + 2G$ scalars, against $2d$ for the
published diagonal.
The input map $B$ is shared and does not change.
That is about four times fewer scalars: $258$ against $1024$ at width $512$ with one group, and
$324$ against $1280$ at the $60$M width with two groups.

\cutsectionup
\section{Experimental Details}
\label{app:repro}
\cutsectiondown

\cutsubsectionup
\subsection{Compute Resources}
\cutsubsectiondown

All runs were performed on NVIDIA H100 GPUs.
A single group-word run at $T_{\max}{=}32$ with the length curriculum takes roughly $2.5$ GPU-hours,
and the group-word study is the bulk of the cost because every cell in \Cref{tab:a5},
\Cref{tab:s5}, \Cref{tab:diaggroups} and \Cref{tab:chart} is ten seeds.
Including the language models, the evaluation sweeps, and failed or exploratory runs, the project
cost approximately $4000$ H100 GPU-hours in total.
The evaluations in \Cref{sec:exp-lm,sec:exp-geom} load existing checkpoints and train nothing; the Toeplitz
measurement is negligible, since it operates on a single $d \times d$ matrix on CPU.

\cutsubsectionup
\subsection{Group Word Problems}
\cutsubsectiondown

The model reads a sequence of generators and emits every prefix product.
Sequences are sampled rather than read from a fixed split, so train and evaluation words do not
overlap by construction.
The recurrence count is drawn from a truncated Poisson with the support top $T_{\max}$ stated in each
caption, following \citet{parcae}.
The length curriculum presents lengths $2, 4, 8, \dots, n$ in equal slices, which
\citet{merrill-logdepth} report is required to learn long $A_5$ words at all.
Every method within a table shares this configuration exactly; the injection module is the only
difference.

\begin{table}[h]
  \centering
  \begin{footnotesize}
  \renewcommand{\arraystretch}{1.02}
  \setlength{\tabcolsep}{4pt}
  \begin{tabular}{@{}lc@{\hspace{1.2em}}lc@{}}
    \toprule
    \textbf{Field} & \textbf{Value} & \textbf{Field} & \textbf{Value} \\
    \midrule
    Optimizer        & Adam                   & Width               & $512$ \\
    Learning rate    & $10^{-4}$              & Recurrent block     & $2$ layers, $4$ heads \\
    Schedule         & constant               & Recurrence          & $T \sim \Lambda$, truncated Poisson \\
    Warmup steps     & $300$                  & Word length $n$     & as stated per table \\
    Batch size       & $128$                  & Curriculum          & $2,4,8,\dots,n$ in equal slices \\
    Grad clip        & $1.0$                  & Precision           & float32, TF32 off \\
    Seeds            & $10$                   & Seed handling       & all retained, none dropped \\
    \bottomrule
  \end{tabular}
  \end{footnotesize}
  \caption{Group-word training configuration, identical across methods within a comparison.}
  \label{tab:gwcfg}
\end{table}

\textbf{Scoring.}
Mean token accuracy averages over prefixes and is high and nearly identical for every injection,
because early prefixes resolve within the first few passes.
Last-token accuracy scores only the full product.
The $T{=}16$ and $T{=}32$ columns of \Cref{tab:a5}, \Cref{tab:s5} and \Cref{tab:diaggroups} evaluate
the $T_{\max}{=}32$ checkpoints in budget; $T{=}48$ re-evaluates the same checkpoints without
retraining.

\textbf{Grouped-diagonal control.}
The control in \Cref{tab:diaggroups} keeps \eqref{eq:diag} elementwise and replaces its $d$ free
rates by $G$ rates repeated across coordinate-aligned blocks of equal width, matching the partition
used by the conformal method at the same $G$.
No rotation is applied.
Initialization, optimizer and schedule are those of the baseline.

\textbf{Charts.}
The three writings compared in \Cref{tab:chart} all produce an exactly orthogonal factor and differ
only in which orthogonal matrices they can reach.
The pairwise chart is \eqref{eq:chart}, a product of $d/2$ planar rotations in fixed coordinate
planes.
The Cayley map sends a learned skew-symmetric $S$ to $(I-S)(I+S)^{-1}$ and covers $SO(d)$ up to a
null set.
The Householder writing takes the $Q$ factor of a QR decomposition of a learned dense matrix and
reaches both components of $O(d)$.
The spectrum is matched across the four rows, so the only thing that varies is the reachable set.

\cutsubsectionup
\subsection{Language Models}
\cutsubsectiondown

We adopt the training scenario of the Parcae codebase \citep{parcae} without modification - the
tokenizer, the data pipeline, the optimizer, the learning-rate schedule, the prelude and coda, and
the recurrence sampler - which trains at approximately $80$ tokens per parameter.
The only edit is the body of the injection module, so the two methods of every comparison differ in the
$d \times d$ map applied between passes.

\textbf{Group-count sweep (\Cref{tab:lmgroups}).}
$60$M parameters on FineWeb-Edu at sequence length $2048$ and global batch $524$k tokens, with the
recurrence count drawn from a truncated Poisson at mean $8$ and validation perplexity measured at
$T{=}8$.
Group counts must divide the $320$ rotation planes at this width, which is what fixes the values of
$G$ that appear in the table.
Truncated BPTT at $4$ is the published recipe; full BPTT at $8$ backpropagates through every pass.
We report both for the diagonal and for $G{=}1$ to show that the choice is worth a small constant and
does not reorder anything, and compare the remaining rows at full BPTT.

\textbf{Geometry and language-model family (\Cref{tab:toeplitz}, \Cref{tab:lm}).}
Three widths, each trained once with each transition operator, giving the six checkpoints on which the Toeplitz
defect is measured.
The recurrent state width equals the model width.
The conformal method uses $G{=}2$ and the coupled angle $\theta_i = \omega_i\,\mathrm{softplus}(\delta)$
of \eqref{eq:chart}.
Recurrence counts are sampled during training around a mean of $8$, and evaluation overrides $T$ at
load time, so the $T \in \{1,8,16,32,48\}$ columns of \Cref{tab:lm} are a test-time depth sweep in
which $T \ge 16$ is beyond the training distribution.

\begin{table}[h]
  \centering
  \begin{footnotesize}
  \renewcommand{\arraystretch}{1.02}
  \setlength{\tabcolsep}{5pt}
  \begin{tabular}{@{}lccc@{}}
    \toprule
    \textbf{Field} & $60$M & $140$M & $370$M \\
    \midrule
    Width $d$                & $640$ & $768$ & $1024$ \\
    State width              & $640$ & $768$ & $1024$ \\
    Mean recurrence (train)  & $8$   & $8$   & $8$ \\
    Injection groups $G$     & $2$   & $2$   & $2$ \\
    Angle mode               & coupled ($\theta = \omega\Delta$) & coupled & coupled \\
    \bottomrule
  \end{tabular}
  \end{footnotesize}
  \caption{Configurations for the three-size family. Everything not listed is the baseline's
  \citep{parcae} and is identical across the two methods at a given width.}
  \label{tab:lmcfg}
\end{table}

\cutsubsectionup

\textbf{Toeplitz defect.}
$\Ab$ is assembled in float64 on CPU, its powers are taken directly, and \eqref{eq:defect} is solved
in closed form by averaging each diagonal of $G_u$.
Probes are drawn uniformly on the unit sphere with a fixed seed shared across methods, so the two methods
of a comparison see the same $256$ directions.

\cutsectionup
\section{Preliminary Group Count}
\label{app:lmgroups}
\cutsectiondown

The sweep that selects $G$ for the language-model comparison is a $60$M model trained at $20$
tokens per parameter, the Chinchilla allocation \citep{hoffmann2022training}.
Full backpropagation through the eight recurrent steps improves perplexity over truncation at four
steps, on both the diagonal ($22.71$ to $22.45$) and $G{=}1$ ($19.64$ to $19.42$), and it does not
change the ordering.
At full backpropagation the diagonal reaches $22.45$, $G{=}2$ reaches $19.08$, $G{=}1$ reaches
$19.42$, and every other count falls between $19.31$ and $19.66$.
Two groups is the value used in \Cref{tab:lm}.

\begin{table}[H]
  \centering
  \begin{small}
  \setlength{\tabcolsep}{6pt}
  \renewcommand{\arraystretch}{1.08}
  \caption{
  \textbf{Preliminary group count on a small language model.}
  Validation perplexity at $T{=}8$ for a $60$M looped model trained at $20$ tokens per parameter
  on FineWeb-Edu, sequence length $2048$ and global batch $524$k tokens, with the recurrence count drawn
  from a truncated Poisson at mean $8$.
  The injection is the only difference between rows.
  The $G{=}2$ entry is the $60$M, $T{=}8$ cell of \Cref{tab:lm}.
  $G{=}320$ is one group per rotation plane, the finest chart this width affords.
  BPTT is how far gradients travel through the loop: truncated $(4)$ is the published recipe and full
  $(8)$ backpropagates through every pass, the comparison of \citet{schwethelm2026much};
  it is a small consistent gain on every matched pair and does not change the ordering, so the
  comparison we quote is the pair at full BPTT.
  The arrow is the relative improvement of $G{=}2$ over the diagonal at full BPTT.
  \boldresultsnote{} Best values are highlighted in \textcolor{LightBlue}{blue}.
  }
  \label{tab:lmgroups}
  \begin{tabular}{@{}llcc@{}}
    \toprule
    Injection & BPTT & $G$ & Validation ppl $\downarrow$ \\
    \midrule
    \multirow[c]{2}{*}{Diagonal} & Truncated $(4)$ & - & $22.71$ \\
      & Full $(8)$ & - & $22.45$ \\
    \midrule
    \multirow[c]{8}{*}{\oursconf}
      & Truncated $(4)$ & $1$ & $19.64$ \\
      & Full $(8)$ & $1$ & $19.42$ \\
      & Full $(8)$ & $2$ & \abgain{19.08}{\gaindown{15.0}} \\
      & Full $(8)$ & $4$ & $19.31$ \\
      & Full $(8)$ & $8$ & $19.48$ \\
      & Full $(8)$ & $64$ & $19.62$ \\
      & Full $(8)$ & $160$ & $19.66$ \\
      & Full $(8)$ & $320$ & $19.52$ \\
    \bottomrule
  \end{tabular}
  \end{small}
\end{table}

\end{document}